# Neuromorphic Sensing for Yawn Detection in Driver Drowsiness

Paul Kielty[1], Mehdi Sefidgar Dilmaghani[1], Cian Ryan[2],

Joe Lemley[2], Peter Corcoran[1]

[1] School of Engineering, University of Galway, Ireland;
[2] Sensing Team, Xperi, Galway, Ireland;

## ABSTRACT

Driver monitoring systems (DMS) are a key component of vehicular safety and essential for the transition from semi-autonomous to fully autonomous driving. A key task for DMS is to ascertain the cognitive state of a driver and to determine their level of tiredness. Neuromorphic vision systems, based on event camera technology, provide advanced sensing of facial characteristics, in particular the behavior of a driver's eyes. This research explores the potential to extend neuromorphic sensing techniques to analyze the entire facial region, detecting yawning behaviors that give a complimentary indicator of tiredness. A neuromorphic dataset is constructed from 952 video clips (481 yawns, 471 not-yawns) captured with an RGB colour camera, with 37 subjects. A total of 95,200 neuromorphic image frames are generated from this video data using a video-to-event converter. From these data 21 subjects were selected to provide a training dataset, 8 subjects were used for validation data, and the remaining 8 subjects were reserved for an 'unseen' test dataset. An additional 12,300 frames were generated from event simulations of a public dataset to test against other methods. A CNN with self-attention and a recurrent head was designed, trained, and tested with these data. Respective precision and recall scores of 95.9% and 94.7% were achieved on our test set, and 89.9% and 91% on the simulated public test set, demonstrating the feasibility to add yawn detection as a sensing component of a neuromorphic DMS.

## 1. INTRODUCTION

Neuromorphic sensing describes a class of technologies designed to mimic biological sensory and perceptual functions. One such device in the field of neuromorphic vision is an event camera. Instead of using a conventional shutter-based approach to capture an image, an event camera asynchronously streams 'events' whenever a pixel detects a relative change in brightness. Each event has 4 parameters: A timestamp, an x and y coordinate corresponding to the location of the pixel that reported the event, and a polarity to denote whether an increase or decrease in brightness triggered the event. Since events are only produced by motion or lighting changes in an otherwise still scene, the output data contains minimal redundant information of components that haven't changed over time. The fundamental shift in operation grants event cameras higher dynamic range and higher temporal resolution than most conventional shutter cameras [1]. Events are logged with microsecond precision and can reach equivalent framerates exceeding 10,000 frames per second. These features make event cameras extremely practical sensors for the analysis of motion.

Drowsy driving is one of the leading causes of motor accidents globally, increasing a driver's risk of an accident by a factor of 5 or more compared to when they are alert [2]. Moreover, in the context of autonomous driving, drowsiness detection for the driver is important as it provides a key indicator that the DMS should present warnings and reduce the dependence of the vehicular control systems on the driver. Many existing drowsiness detection systems tend to focus on analysis of eye-regions and the blinking behaviour of the subject [3]. While such analysis is essential it can be supplemented with other physiological signs that a driver is feeling weary or gives indications of tiredness. One very useful indicator is that of yawning – a noticeable facial action that should be clearly distinguishable by a neuromorphic vision system and offer a useful complimentary indicator to the more conventional eye-blink analysis.

A research area of increasing popularity is the monitoring of a human subject with a view to detect certain facial or body actions that provide information about the cognitive or physiological state. Monitoring these states requires precise observation of the subject and in applications dealing with rapid movements such as blinks, eye motion, and micro-expressions, performance is often 'bottlenecked' by the hardware limitations of standard imaging technologies. The advent of neuromorphic vision sensors promises a new era for these systems by addressing a variety of these limitations, including framerate, power consumption, and low-light performance. In this paper we utilize an event camera's ability to isolate moving features in combination with a recurrent convolutional neural network (CNN) to detect yawns, a valuable signal in drowsiness detection. The study of neuromorphic vision for driver monitoring and drowsiness detection is found in literature, but to the best of our knowledge this paper is the first to directly classify yawns from event data.

## 2. LITERATURE REVIEW

### 2.1 Overview of driver drowsiness detection methods

To detect driver drowsiness, various methods have been proposed in recent years which can be categorized in three main groups: The first group checks the driving pattern of the vehicle by observing the accelerator, breaks, steering wheel etc. The second group uses physiological signals of the driver and uses tools such as EEG, ECG, and EOG. The last group belongs to computer vision-based schemes and attempts to analyze the drivers' eyes, hands, and various body movements [4]. The first group is based on traditional driver behavior analysis methods and depends highly on the roads, weather, driver, and other factors. The second group has shown some good results for academic research. However, they use intrusive approaches requiring probes connected to the driver's body which are impractical in real driving situations. The third group has none of the above-mentioned shortcomings, and due to recent developments in computer vision algorithms, sensors, and camera technologies, is where we believe the best practical solutions can be found.

### 2.2 Driver drowsiness & yawns with visible-light imaging

The authors of [2] propose a scheme which checks the openness of the eyes and mouth to determine drowsiness. This is achieved by detecting facial landmarks on visible light frames, and then measuring the distance between chosen sets of landmarks. If distance between the lips remains above some threshold for 30 frames, it is considered a yawn. In [5], Abtahi et al. detect yawns with a series of prerequisite detection steps. First is the face is detected, then the mouth and eye locations. The eye region is used as a reference for verifying the mouth position and openness. In [6] a comprehensive method of fatigue detection is proposed which checks blinking, head state, frequency of nods, and yawns together. The inclusion of other parameters is useful, however, their analysis methods are based on traditional image processing and they also depend on extraction the mouth area, with comparisons consecutive frames, to detect yawns. The reliance of measurement of the mouth in these methods will see many yawns missed in practice, as some people reflexively cover their mouth with a hand yawning. In addition, the mouth can remain open for other purposes such as talking, laughing, or sighing, which could lead to many false positives. Yawns can manifest in many forms, such as the tensing of various facial muscles other than the mouth, or accompanied by stretches of the upper body and arms. For a robust yawn detector, more of these features should be monitored. For their yawn detection CNN, Zhang et al. [7] discuss edge detection as a pre-processing technique to reduce complexity while preserving the yawning information of an RGB image. Event cameras provide similar benefits in motion identifying tasks, as events generated by movement typically appear on the edges of the moving object. This feature is inherent to the design of an event camera, so we believe it can provide a more robust signal for yawn detection.

### 2.3 Driver monitoring with event cameras

Real-time driver monitoring deals with watching blinks, different types of eye movements such as saccades, breathing and other changes in drivers' condition. Some of these movements are too fast to be reliably caught by conventional cameras with limited framerates. Until recent years, this was a common problem across most driver monitoring research, and is one of many challenges that makes neuromorphic vision sensors a promising basis for driver monitoring systems, as demonstrated by Ryan et al. with their work on real-time blink detection and eye tracking using event cameras [3]. Chen et al propose an event camera system for the detection of driver drowsiness by performing a two-stage filtering process on the event stream [8]. The goal of this filtering is to remove any events that are not in the eye or mouth regions. They then infer drowsiness from the isolated eye and mouth movements, though they do not detect yawns specifically. This system requires the driver's head to be held quite still for the filter to isolate the eye and mouth regions correctly. Our method is more flexible, using a recurrent CNN algorithm which learns the relevant features to extract from anywhere within the frame. Our datasets, pre-processing, network design and training are detailed in section III. Section IV outlines our results and how our method compares to others in literature. Finally, in section V we share our conclusions and discussions of future work.

# 3. METHODOLOGY

## 3.1 Our Dataset

A frequently encountered barrier in neuromorphic vision research is lack of large-scale public datasets, and yawning data is no exception. This lack of data has led to the development of event simulators such as v2e [9], which enables the generation of realistic synthetic events from RGB frames. To generate a synthetic event yawn dataset, we took RGB frames from a private Xperi dataset focused on driver drowsiness. These frames were captured by a Logitech BRIO 4k camera positioned behind the steering wheel of a driving simulator. Subjects were recorded driving in this simulator at 5PM, 2AM, and 5AM local time, each recording 1 hour long. They were required to not consume any stimulants 12 hours before or during the acquisition, and sleeping was not permitted between any of the recording sessions. The audio of each session was also recorded and later annotated by a team within Xperi. These annotations contained the start time and duration of all yawns.

The RGB frames are all timestamped so these yawn audio timestamps can be transferred to label each frame. However, simply aligning the two sets of timestamps for frame-wise labels has a significant flaw. There are various ways a person can yawn, many of which are audible at different times. For example, some are mostly silent when the mouth opens and is held open, followed by a loud exhale as the mouth closes. The inverse can also occur, or it could be consistently audible from start to finish. This results in an irregular misalignment between the visible cues and the audible cues of a yawn. Considering this, the network and dataset created for our yawn detector assigns a single class to a sequence of frames, to denote if a yawn occurred anywhere within that sequence.

The yawn audio labels have a mean duration of 4.03s with a standard deviation of 2.26s and a maximum of 9.63s. The sequence duration chosen for the network to classify was 10s. To create our RGB video yawn dataset, each yawn sample is temporally centered in a frame sequence spanning 10s. This extended duration reduces the likelihood of a yawn being clipped by the misalignment of the audible and visible yawn components. The video was specified to be collected at 30 frames per second, giving an expected 300 RGB frames in each sequence. However, the framerate was typically lower in the AM sessions due to an increased exposure time. For this reason, the frame sequences were created by timestamp instead of taking a fixed 300 frames. In the final set of RGB yawn sequences, 48.6% had fewer than 300 frames. The frame counts of these shorter sequences only have a mean and standard deviation of 203.63 frames and 48.52 frames respectively. To maintain accuracy of the temporal information of the synthetic events despite the missing frames, each frame's timestamp is also passed to v2e at simulation time. Events are simulated from the differences between consecutive pairs of frames, and the timestamps of these events are distributed over the time span of the source frame pair. Before simulation, all RGB frames were cropped to a 500x500 area containing the face to reduce the simulation time and the size of the output event data. A set of non-yawn sequences was created by adding a 10s offset to the end of each yawn sequence and saving another 10s of frames, provided that these frames did not collide with a subsequent yawn sequence.

## 3.2 Pre-processing of Event Camera Data

To input to a CNN, the information must first be transformed into a 2D array. This is most commonly achieved by accumulating a group of events and summing the positive and negative events at each pixel location to create a 2D frame [1]. The events can be grouped by a fixed event count, such that each frame is created from the same number of events, or grouped over a fixed duration. Using a fixed event count per frame provides a level of assurance that each frame has some minimum amount of spatial information, at the loss of much temporal information. We hypothesize that the second approach of accumulating events over a fixed duration is more applicable for yawn detection. This yields frames at fixed rate, much like a conventional camera's output, and the temporal information carried in a sequence of these frames can be useful when identifying yawns from other actions such as speech, due to differences in the rate of mouth motion. The choice of this event frame duration should be informed by the requirements of the underlying task. Accumulating events over a long period risks an aliasing effect, where speech frames could appear as one long mouth open sequence if insufficiently sampled. On the other hand, using too short a period can yield many frames with low spatial information. For our dataset, each frame is generated by accumulating events over a duration of 0.1s, resulting in frame sequences of 100 frames at 10FPS. This reduction from the 30FPS of the source data has 3 primary justifications: (1) A higher frame frequency is unnecessary to distinguish a yawn from speech. (2) With fewer than 300 frames in many RGB sequences, accumulating an equal or greater number of event frames would require many event frames between consecutive RGB frames. This can create a freezing effect in the event videos due to several frames showing the same motion. (3) A reduction from 300 to 100 frames for each sample carries a significant speedup to network training. The event frames' pixel values are clipped to +/-10 and then normalized between 0,255. Sample event frames from a yawn sequence are shown in Fig. 1.

Figure 1. Sample event frames of a yawn sequence.

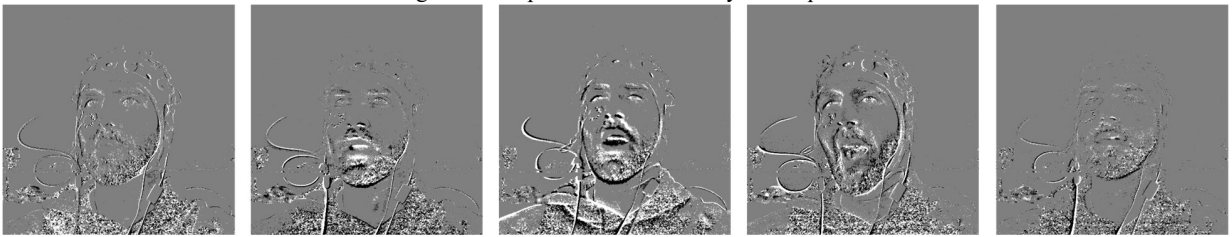

### 3.3 YawDD Dataset

To assess our yawn detector on a public dataset, events were simulated from the YawDD dataset [10]. This dataset is comprised of videos taken from the dashboard and rear-view mirror of a car. Our proposed system uses a camera behind the steering wheel, so the camera position of the dashboard videos is much more appropriate for our tests. As such, the rear-view mirror videos have not yet been included for experiments.

The 30FPS RGB videos were simulated to 10FPS event videos identically to our custom dataset. The dashboard videos were annotated with the start and stop frames of yawns and 100 frame sequences were extracted with these yawns centered. Non-yawn sequences were also saved from the frames between yawns. This totalled to 12,300 synthetic event frames, containing 78 yawn sequences and 45 non-yawn sequences.

### 3.4 Network Architecture

The most commonly monitored feature for yawn detection is the mouth openness, but methods that rely too heavily on this have significant weaknesses, such as the risk of false positive predictions when the mouth is open for speech or laughter. To reduce this risk, some methods analyse more facial features or input sequences of video frames instead of single images to use of temporal information for each prediction. An additional flaw - which is extremely challenging to solve in these systems - is not handling the common case where a person reflexively covers their mouth with their hand when yawning. This is reflective of a general oversimplification of how yawns manifest in many of the current approaches. Specific monitoring of the mouth and eyes is the basis of the majority of these current methods, with little consideration of other possible cues on the face or body actions that often accompany a yawn, such as the hand over the mouth or large stretches of the upper body and arms. For this reason our proposed solution does not use facial landmarks or other deliberately programmed features to make a prediction. Instead, we rely on CNN components learning the relevant features in a full input image. With the ability to extract yawn-related features from individual frames, a recurrent portion of the network learns how these features change over time in a video of someone yawning. This allows the distinction of yawns from other actions that can appear similar on single frames, such as speech or laughter.

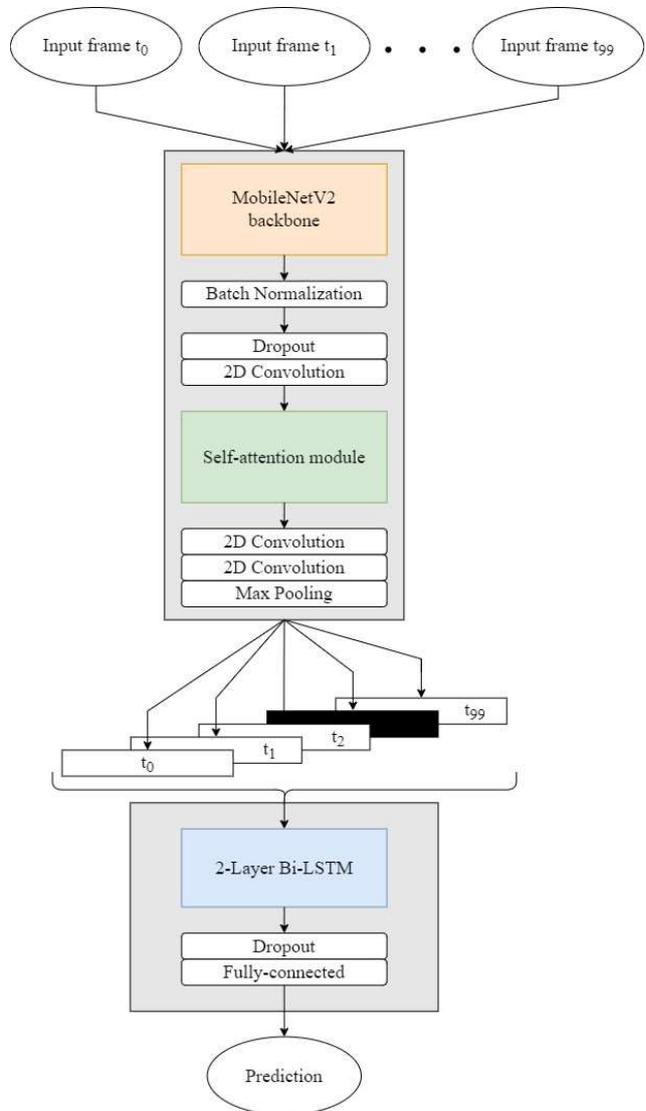

Figure 2. Yawn detection network architecture.

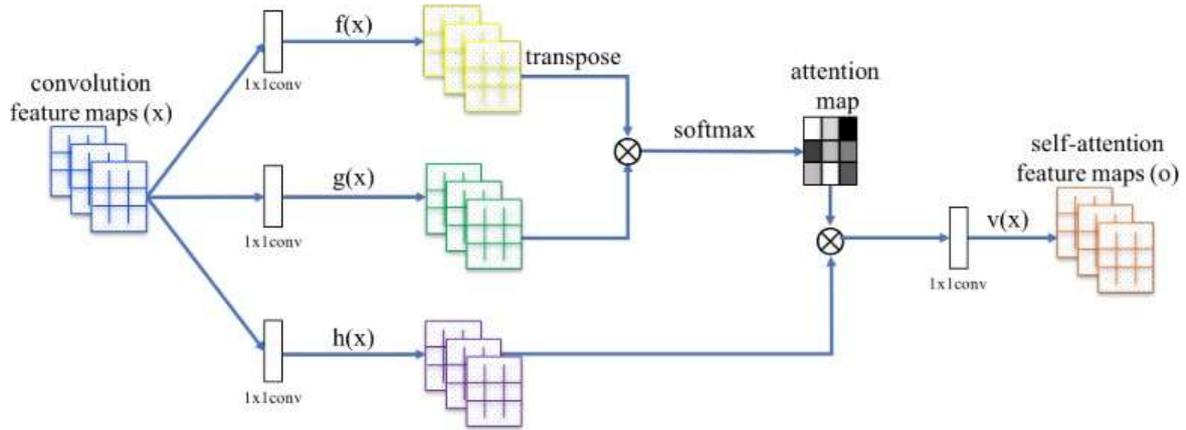

Figure 3. An expanded diagram of the self-attention module in our proposed network. [13]

Fig. 2 gives a high-level overview of the model architecture. The MobileNetV2 network is used for feature extraction of the input frames. In their paper, Sandler et al. ex [11] demonstrate the impressive performance of MobileNetV2 as a feature extractor with an efficient, lightweight architecture. The small size made this an appealing backbone for our network because learning from a single data sample the feature extraction of 100 frames. The model we used was pretrained on the ImageNet dataset [12].

After this initial feature extraction, batch normalization and channel reduction by 2D convolution are applied to prepare the features for a self-attention module. Recent years have seen self-attention introduced to many CNN tasks for it's ability to contextualize and apply a weighting to input features, with only a small computational cost. The self-attention module in our proposed network is implemented according to [13]. Fig. 3.4 gives the expanded diagram of this module.

When the attended feature maps are generated for every frame of the input sequence, they are stacked and passed to the recurrent head of the network. This is comprised of a 2 stacked bi-directional LSTM layers [14]. By it's use of memory cells, the LSTM architecture is has greater ability to relate long temporal sequences than most recurrent structures. This was required for our input sequences of 100 frames. This also grants flexibility for inputs of different lengths, which can prove particularly useful when creating frames from a sequence of event data, as there is a enormous range of possible representations which can massively vary the number of inputs. For each frame sequence, the final output of the LSTM layers is flattened and passed to a fully-connected layer to generate the output prediction. Dropout layers were added to the network after some initial tests showed overfitting on the training data.

### 3.5 Training Details

Our simulated event dataset was split into three sets for training, validation, and testing. The breakdown of each set is shown in table x. There is no overlap of subjects between the three sets. The training sequences were augmented to achieve better generalization. This includes rotating 50% of sequences within ±10°, mirroring about a vertical axis, and cropping to squares of randomized size and position (within some limits to ensure the full face is still visible). The augmentations were only randomized between sequences, so each frame in a sequence had identical transformations applied.

All frames were downsampled to 256x256 using pixel area relation before input to the network. The network was trained for 100 epochs with a batch size of 5. The initial learning rate of $1 \times 10^{-4}$ was halved every 10 epochs. Binary cross entropy loss was calculated between the predicted and actual labels of each sequence in the validation set.

Table 1. Distribution of our event yawn dataset partitions.

|  | Train set | Valid set | Test set | Total |
|---|---|---|---|---|
| Subject counts | 21 | 8 | 8 | 37 |
| Event frame counts | 65,200 | 15,000 | 15,000 | 95,200 |
| Yawn sequence counts | 331 | 75 | 75 | 481 |
| Non-yawn sequence counts | 321 | 75 | 75 | 471 |

## 4. RESULTS

The 10 epochs with the lowest validation loss were tested on (a) our test set and (b) the simulated YawDD dash set. The best model with determined by the highest mean F1 score on both sets. The precision, recall, and F1 score of this model on each dataset partition are listed in table 2. The confusion matrices in table 3 detail the per-class performance on both test sets. Average inference time was measured at 0.44s per 100 frame sequence, where each sequence corresponds to 10s of data, granting a large margin for real-time inference.

Table 2. Results of our best model tested on all of our synthetic event sets.

| Dataset | Precision | Recall | F1 |
|---|---|---|---|
| Train | 97.0% | 98.1% | 97.6% |
| Valid | 90.4% | 100% | 94.9% |
| Test (ours) | 95.9% | 94.7% | 95.3% |
| YawDD | 89.9% | 91.0% | 90.4% |

Table 3. Confusion matrices of model predictions on (a) our test set and (b) YawDD dash set.

(a) Our test set

|  |  | Actual values | |
|---|---|---|---|
|  |  | Yawn | Non-yawn |
| Predicted values | Yawn | 71 | 3 |
|  | Non-yawn | 4 | 72 |

(b) YawDD dash set

|  |  | Actual values | |
|---|---|---|---|
|  |  | Yawn | Non-yawn |
| Predicted values | Yawn | 71 | 8 |
|  | Non-yawn | 7 | 37 |

Our method is compared to several other yawn detection methods in table 4 by their results on the YawDD dataset. We have achieved a high level of performance, surpassing these methods with no YawDD data present in the training or validation sets.

Table 4. Comparison of yawn detection methods by performance on YawDD dataset.

| Method | Precision | Recall | F1 |
|---|---|---|---|
| Dong et al [15] | 100% | 67% | 80.2% |
| Zhang et al. [7] | 87.1% | 88.6% | 87.8% |
| Akrout and Mahdi [16] | 82% | 83% | 82.5% |
| Omidyeganeh et al. [17] | 70% | 70% | 70% |
| **Ours** | 89.9% | 91% | **90.4%** |

## 5. CONCLUSION

This paper presents a proof of concept of a yawn detection system using event cameras. We outline the creation of a synthetic event dataset and the processing of the raw events for compatibility with existing CNN architectures. A lightweight network was built with initial feature extraction layers, self-attention and a recurrent head for event video classification. After training, the model exhibited impressive performance on our testing dataset with high speeds. In addition, events were simulated from a public YawDD dataset. The results on this dataset demonstrate our yawn detector method generalizes to new data, outperforming others in literature. Our algorithm benefits from a more robust design than the most common approaches which require precise positioning of the face or accurate facial landmarks, which isn't possible for many yawns in practice e.g. when the mouth is covered by a hand. With promising results and low requirements we can justify the inclusion of a yawn detection algorithm in event-based driver drowsiness detection systems.

There remains much scope for future research. Currently the most significant absence in our testing is the lack of real event data. Further data acquisitions for driver monitoring tasks are planned for the near future with an event camera as one of the sensors. Additional public yawn datasets will also be converted to synthetic events.

## ACKNOWLEDGMENTS

This research was conducted with the financial support of Science Foundation Ireland at ADAPT, the SFI Research Centre for AI-Driven Digital Content Technology at the University of Galway [13/RC/2106_P2]. For the purpose of Open Access, the author has applied a CC BY public copyright licence to any Author Accepted Manuscript version arising from this submission.

# REFERENCES

[1] G. Gallego, T. Delbruck, G. Orchard, C. Bartolozzi, B. Taba, A. Censi, S. Leutenegger, A. J. Davison, J. Conradt, K. Daniilidis, and D. Scaramuzza, "Event-based vision: A survey," IEEE Transactions on Pattern Analysis and Machine Intelligence **44**, 154–180 (jan 2022).

[2] S. Kumari, K. Akanksha, S. Pahadsingh, Swati, and S. Singh, "Drowsiness and yawn detection system using python," in Proceedings of International Conference on Communication, Circuits, and Systems, S. K. Sabut, A. K. Ray, B. Pati, and U. R. Acharya, eds., 225–232, Springer Singapore, Singapore (2021).

[3] C. Ryan, B. O'Sullivan, A. Elrasad, A. Cahill, J. Lemley, P. Kielty, C. Posch, and E. Perot, "Real-time face & eye tracking and blink detection using event cameras," Neural Networks **141**, 87–97 (2021).

[4] F. Hasan and A. Kashevnik, "State-of-the-art analysis of modern drowsiness detection algorithms based on computer vision," in 2021 29th Conference of Open Innovations Association (FRUCT), 141–149 (2021).

[5] S. Abtahi, B. Hariri, and S. Shirmohammadi, "Driver drowsiness monitoring based on yawning detection," in 2011 IEEE International Instrumentation and Measurement Technology Conference, 1–4 (2011).

[6] D. Liu, C. Zhang, Q. Zhang, and Q. Kong, "Design and implementation of multimodal fatigue detection system combining eye and yawn information," in 2020 IEEE 5th International Conference on Signal and Image Processing (ICSIP), 65–69 (2020).

[7] W. Zhang and J. Su, "Driver yawning detection based on long short term memory networks," in 2017 IEEE Symposium Series on Computational Intelligence (SSCI), 1–5 (2017).

[8] G. Chen, L. Hong, J. Dong, P. Liu, J. Conradt, and A. Knoll, "Eddd: Event-based drowsiness driving detection through facial motion analysis with neuromorphic vision sensor," IEEE Sensors Journal **20**(11), 6170–6181 (2020).

[9] Y. Hu, S.-C. Liu, and T. Delbruck, "v2e: From video frames to realistic dvs events," in 2021 IEEE/CVF Conference on Computer Vision and Pattern Recognition Workshops (CVPRW), 1312–1321 (2021).

[10] S. Abtahi, M. Omidyeganeh, S. Shirmohammadi, and B. Hariri, "Yawdd: A yawning detection dataset," in Proceedings of the 5th ACM Multimedia Systems Conference, MMSys '14, 24–28, Association for Computing Machinery, New York, NY, USA (2014).

[11] M. Sandler, A. Howard, M. Zhu, A. Zhmoginov, and L.-C. Chen, "Mobilenetv2: Inverted residuals and linear bottlenecks," in 2018 IEEE/CVF Conference on Computer Vision and Pattern Recognition, 4510–4520 (2018).

[12] O. Russakovsky, J. Deng, H. Su, J. Krause, S. Satheesh, S. Ma, Z. Huang, A. Karpathy, A. Khosla, M. Bernstein, A. C. Berg, and L. Fei-Fei, "ImageNet Large Scale Visual Recognition Challenge," International Journal of Computer Vision (IJCV) **115**(3), 211–252 (2015).

[13] H. Zhang, I. Goodfellow, D. Metaxas, and A. Odena, "Self-attention generative adversarial networks," (2018).

[14] S. Hochreiter and J. Schmidhuber, "Long short-term memory," Neural computation **9**, 1735–80 (12 1997).

[15] B.-T. Dong, H.-Y. Lin, and C.-C. Chang, "Driver fatigue and distracted driving detection using random forest and convolutional neural network," Applied Sciences **12**(17) (2022).

[16] B. Akrout and W. Mahdi, "Yawning detection by the analysis of variational descriptor for monitoring driver drowsiness," in 2016 International Image Processing, Applications and Systems (IPAS), 1–5 (2016).

[17] M. Omidyeganeh, S. Shirmohammadi, S. Abtahi, A. Khurshid, M. Farhan, J. Scharcanski, B. Hariri, D. Laroche, and L. Martel, "Yawning detection using embedded smart cameras," IEEE Transactions on Instrumentation and Measurement **65**(3), 570–582 (2016).